\documentclass[conference]{IEEEtran}
\usepackage{cite}
\usepackage{graphicx}
\usepackage{latexsym}
\usepackage{amssymb}
\usepackage{url}
\usepackage[fleqn]{amsmath}
\usepackage[varg]{txfonts}
\usepackage{algpseudocode}
\usepackage{refcount}
\usepackage{footnote}
\usepackage{breqn}
\usepackage{hhline}
\usepackage{multirow}

\makesavenoteenv{tabular}
\makesavenoteenv{table}

\ifCLASSINFOpdf
\else
\fi
\begin{document}
%
\title{Compressing Recurrent Neural Network\\ with Tensor Train}

\author{\IEEEauthorblockN{Andros Tjandra,
Sakriani Sakti, Satoshi Nakamura}

\IEEEauthorblockA{Graduate School of Information Science, Nara Institute of Science and Technology, Japan\\
Email : andros.tjandra.ai6@is.naist.jp, ssakti@is.naist.jp, s-nakamura@is.naist.jp}
}


%


\maketitle

\begin{abstract}
Recurrent Neural Network (RNN) are a popular choice for modeling temporal and sequential tasks and achieve many state-of-the-art performance on various complex problems. However, most of the state-of-the-art RNNs have millions of parameters and require many computational resources for training and predicting new data. This paper proposes an alternative RNN model to reduce the number of parameters significantly by representing the weight parameters based on Tensor Train (TT) format. In this paper, we implement the TT-format representation for several RNN architectures such as simple RNN and Gated Recurrent Unit (GRU). We compare and evaluate our proposed RNN model with uncompressed RNN model on sequence classification and sequence prediction tasks. Our proposed RNNs with TT-format are able to preserve the performance while reducing the number of RNN parameters significantly up to 40 times smaller.
\end{abstract}


%
\IEEEpeerreviewmaketitle

\section{Introduction}

Temporal and sequential modeling are important subjects in machine learning. RNNs architecture has recently become a popular choice for modeling temporal and sequential tasks. Although RNNs have been researched for about two decades \cite{elman1990finding, hochreiter1997long}, their recent resurgence reflects improvements in computer hardware and the growth of available datasets. Many state-of-the-arts in speech recognition \cite{hannun2014deep, amodei2015deep} and machine translation \cite{wu2016google, bahdanau2014neural, sutskever2014sequence} has been achieved by RNNs. 

However, most RNN models are computationally expensive and have a huge number of parameters. Since RNNs are constructed by multiple linear transformations followed by nonlinear transformations, we need multiple high-dimensional dense matrices as parameters. In time-steps, we need to apply multiple linear transformations between our dense matrix with high-dimensional input and previous hidden states. Especially for state-of-the-art models on speech recognition \cite{amodei2015deep} and machine translation \cite{wu2016google}, such huge models can only be implemented in high-end cluster environments because they need massive computation power and millions of parameters. This limitation  hinders the creation of efficient RNN models that are fast enough for massive real-time inference or small enough to be implemented in low-end devices like mobile phones \cite{schuster2010speech} or embedded systems with limited memory.

To bridge the gap between high-performance state-of-the-art model with efficient computational and memory costs, there is a trade-off between high accuracy model and fast efficient model. A number of researchers have done notable work to minimize the accuracy loss and maximize the model efficiency. Hinton et al. \cite{hinton2015distilling} and Ba et al. \cite{ba2014deep} successfully compressed a large deep neural network into a smaller neural network by training the latter on the transformed softmax outputs from the former. Distilling knowledge from larger neural networks has also been successfully applied to recurrent neural network architecture by \cite{tang2016recurrent}. Denil et al.\cite{denil2013predicting} utilized low-rank matrix decomposition of the weight matrices. A recent study by Novikov et al.\cite{novikov2015tensorizing} replaced the dense weight matrices with Tensor Train (TT) format \cite{oseledets2011tt} inside convolutional neural network (CNN) model. With the TT-format, they significantly compress the number of parameters and kept the model accuracy degradation to a minimum. However, to the best of our knowledge, no study has focused on compressing more complex neural networks such as RNNs with tensor-based representation.

In this work, we propose TT-RNN, which is an RNN architecture based on TT-format. We apply TT-format to reformulate two different RNNs: a simple RNN and a GRU RNN. Our proposed RNN architectures are evaluated using two different tasks: sequence classification and sequence prediction. In Section \ref{sec:background}, we briefly review RNN. In Section \ref{sec:proposed}, we describe the details of our proposed TT-RNN architecture. In Section \ref{sec:expr}, we describe the tasks and datasets, followed by the experimental results. We present related works in Section \ref{sec:relatedwork}. Finally, we conclude our result in Section \ref{sec:conclusion}.

\section{Recurrent Neural Network} \label{sec:background}
\subsection{Simple Recurrent Neural Network} \label{sec:rnn}
\begin{figure}[h]
    \centering
    \includegraphics[width=4.5cm]{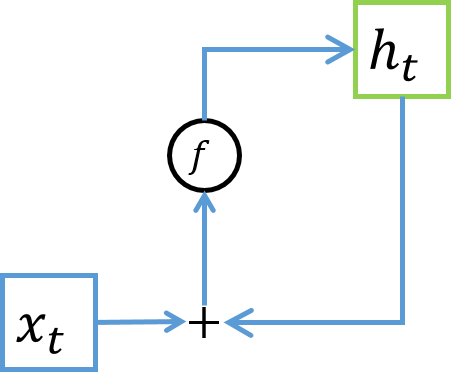}
    \caption{Recurrent Neural Network} 
    \label{fig:rnn}
\end{figure}

An RNN is a kind of neural network architecture that models sequential and temporal dependencies \cite{graves2013speech}. Typically, we define input sequence $\mathbf{x}=(x_1,...,x_{T})$, hidden vector sequence $\mathbf{h}=(h_1,...,h_{T})$ and output vector sequence $\mathbf{y}=(y_1,...,y_T)$. As illustrated in Fig.~\ref{fig:rnn}, a simple RNN at time $t$ is can be formulated as:
\begin{eqnarray} \label{eq:simplernn}
h_t &=& f(W_{xh}x_t + W_{hh}h_{t-1}  + b_h) \\
y_t &=& g(W_{hy}h_t  + b_y). 
\end{eqnarray} where $W_{xh}$ represents the weight parameters between the input and hidden layer, $W_{hh}$ represents the weight parameters between the hidden and hidden layer, $W_{hy}$ represents the weight parameters between the hidden and output layer, and $b_h$ and $b_y$ represent bias vectors for the hidden and output layers. Functions $f(\cdot)$ and $g(\cdot)$ are nonlinear activation functions, such as sigmoid or tanh.
\subsection{Gated Recurrent Neural Network} \label{sec:gatedrnn}
Simple RNNs cannot easily be used for modeling datasets with long sequences and long-term dependency because the gradient can easily vanish or explode \cite{bengio1994learning, hochreiter2001gradient}. This problem is caused by the effect of bounded activation functions and their derivatives. Therefore, training a simple RNN is more complicated than training a feedforward neural network. Some researches addressed the difficulties of training simple RNNs. For example, Le et al. \cite{le2015simple} replaced the activation function that causes the vanishing gradient with a rectifier linear (ReLU) function. With an unbounded activation function and identity weight initialization, they optimized a simple RNN for long-term dependency modeling. Martens et al. \cite{martens2011learning} used a second-order Hessian-free (HF) optimization method rather than the first-order method such as gradient descent. However, estimation of the second-order gradient requires extra computational steps. Modifying the internal structure from RNN by introducing gating mechanism also helps RNNs solve the vanishing and exploding gradient problems. The additional gating layers control the information flow from the previous states and the current input \cite{hochreiter1997long}. Several versions of gated RNNs have been designed to overcome the weakness of simple RNNs by introducing gating units, such as Long-Short Term Memory (LSTM) RNN and GRU RNN. In the following subsections, we explain both in more detail.

\subsubsection{Long-Short Term Memory RNN} \label{sec:lstm}
\begin{figure}[h]
    \centering
    \includegraphics[width=6.5cm]{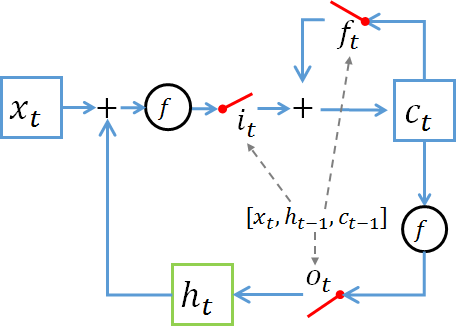}
    \caption{Long Short Term Memory Unit.}
    \label{fig:lstmrnn}
\end{figure}

The LSTM RNN was proposed by Hochreiter et al.\cite{hochreiter1997long}. LSTM is a gated RNN with three gating layers and memory cells, utilizes the gating layers to control the current memory states by retaining the valuable information and forgetting the unneeded information. The memory cells store the internal information across time steps. As illustrated in Fig.~\ref{fig:lstmrnn}, the LSTM hidden layer values at time $t$ are defined by the following equations \cite{graves2013hybrid}:
\begin{eqnarray}
i_t &=& \sigma(W_{xi} x_t + W_{hi} h_{t-1} + W_{ci} c_{t-1} + b_i) \nonumber \\
f_t &=& \sigma(W_{xf} x_t + W_{hf} h_{t-1} + W_{cf} c_{t-1} + b_f) \nonumber \\
c_t &=& f_t \odot c_{t-1} + i_t \odot \tanh(W_{xc} x_t + W_{hc} h_{t-1} + b_c) \nonumber \\
o_t &=& \sigma(W_{xo} x_t + W_{ho} h_{t-1} + W_{co} c_t + b_o) \nonumber \\
h_t &=& o_t \odot \tanh(c_t) \nonumber 
\end{eqnarray}
where $\sigma(\cdot)$ is sigmoid activation function and $i_t, f_t, o_t$ and $c_t$ are respectively the input gates, the forget gates, the output gates and the memory cells. The input gates retain the candidate memory cell values that are useful for the current memory cell and the forget gates retain the previous memory cell values that are useful for the current memory cell. The output gates retain the memory cell values that are useful for the output and the next time-step hidden layer computation. 

\subsubsection{Gated Recurrent Unit RNN} \label{sec:gru}
\begin{figure}[h]
    \centering
    \includegraphics[width=6.5cm]{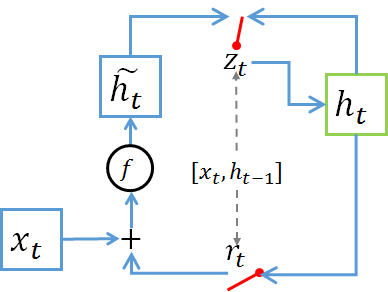}
    \caption{Gated Recurrent Unit}
    \label{fig:grurnn}
\end{figure}

The GRU RNN was proposed by Cho et al. \cite{cho2014learning} as an alternative to LSTM. There are several key differences between GRU and LSTM. First, a GRU does not have memory cells \cite{chung2014empirical}. Second, instead of three gating layers, it only has two: reset gates and update gates.
As illustrated in Fig.~\ref{fig:grurnn}, the GRU hidden layer at time $t$ is defined by the following equations \cite{cho2014learning}:
\begin{eqnarray}
r_t &=& \sigma(W_{xr} x_t + W_{hr} h_{t-1} + b_r) \label{eq:grureset} \\
z_t &=& \sigma(W_{xz} x_t + W_{hz} h_{t-1} + b_z) \label{eq:gruupdate} \\
\tilde{h_t} &=& f(W_{xh} x_t + W_{hh} (r_t \odot h_{t-1}) + b_h) \label{eq:grucandidate}\\
h_t &=& (1 - z_t) \odot h_{t-1} + z_t \odot \tilde{h_t} \label{eq:gruhidden}
\end{eqnarray}
where $\sigma(\cdot)$ is a sigmoid activation function, $f(\cdot)$ is a tanh activation function, $r_t, z_t$ are the reset and update gates, $\tilde{h_t}$ is the candidate hidden layer values, and $h_t$ is the hidden layer values at time-$t$. The reset gates control the previous hidden layer values that are useful for the current candidate hidden layer. The update gates decide whether to keep the previous hidden layer values or replace the current hidden layer values with the candidate hidden layer values. 
GRU can match LSTM's performance and its convergence speed sometimes surpasses LSTM, despite having one fewer gating layer \cite{chung2014empirical}.

In this section, we provided the formulation and the details for several RNNs. As we can see, most of the RNNs consist of many dense matrices that represents a large number of weight parameters that are required to represent all of the RNN models. In the next section, we present an alternative RNN model that significantly reduces the number of parameters and simultaneously preserves the performance.

\section{Proposed Tensor Train based RNN} \label{sec:proposed}
In this section, we describe our proposed approach to compress RNN using Tensor Train (TT) format representation. We start with the description of Tensor Train \cite{oseledets2011tt} and then represent the linear transformation operation in the TT-format \cite{novikov2015tensorizing}. After that, we describe the details of our approach for TT-RNN including a simple RNN and more sophisticated RNN with gating units. Applying the TT-format to represent the weight parameters in RNN presents more difficulties compared to the standard feedforward NN. To tackle this problem, we also propose a local initialization trick in the last subsection. 

\subsection{Tensor Train (TT) format}
Before defining Tensor Train (TT) format, we will explain the notations which we borrow from \cite{oseledets2011tt, novikov2015tensorizing} that will be used in later sections. In general cases, one-dimensional arrays are called vectors, two-dimensional arrays are called matrices, and all higher multidimensional arrays are commonly called tensors. 

We represent vectors with lower case letters (e.g., $b$), matrices with upper case letters (e.g., $W$) and tensors with calligraphic upper case letters (e.g., $\mathcal{W}$). Each element from the vectors, matrices and tensors is represented explicitly using indexing in every dimension. For example, $b(i)$ is the $i$-th element from vector $b$, $W(p,q)$ is the element of the $p$-th row and the $q$-th column from matrix $W$, $\mathcal{W}(j_1,..,j_d)$ is the element at index $(j_1, .., j_d)$ of tensor $\mathcal{W}$ and $d$ is the order of tensor $\mathcal{W}$. Based on previous description  \cite{novikov2015tensorizing}, we assume that $d$-dimensional array (tensor) $\mathcal{W}$ is represented in TT-format\cite{oseledets2011tt} if for each $k \in \{1,..,d\}$ and for each possible value of the $k$-th dimension index $j_k \in \{1,..,n_k\}$ there exists a matrix $G_k[j_k]$ such that all elements of $\mathcal{W}$ can be computed as the following equation :  
\begin{multline} \label{eq:ttprod1}
\mathcal{W}(j_1,j_2,..,j_{d-1},j_d) = \\ G_1[j_1] \cdot G_2[j_2] ...  G_{d-1}[j_{d-1}] \cdot G_{d}[j_{d}].
\end{multline}
For all matrices $G_k[j_k]$ related to the same dimension $k$, they must be represented with size $r_{k-1} \times r_k$, where $r_0$ and $r_d$ must be equal to 1 to retain the final matrix multiplication result as a scalar. 
In TT-format, we define a sequence of rank $\{{r_k}\}_{k=0}^{d}$ and we call them TT-rank from tensor $\mathcal{W}$. The set of matrices $\mathcal{G}_k = \{G_k[j_k]\}_{j_k=1}^{n_k}$ where the matrices are spanned in the same index are called TT-core. 
We can describe Eq.\ref{eq:ttprod1} in detail by enumerating the index $q_{k-1} \in \{1,..,r_{k-1}\}$ and $q_{k} \in \{1,..,r_{k}\}$ in matrix $G_k[j_k]$ across all $k \in \{1,..,d\}$:
\begin{multline} 
\mathcal{W}(j_1,j_2,..,j_{d-1},j_d) = \\ \sum_{q_0,..,q_{d}}G_1[j_1](q_0,q_1) .. G_{d}[j_{d}](q_{d-1},q_d).
\end{multline}
\begin{figure}[h]
    \centering
    \includegraphics[width=8.5cm]{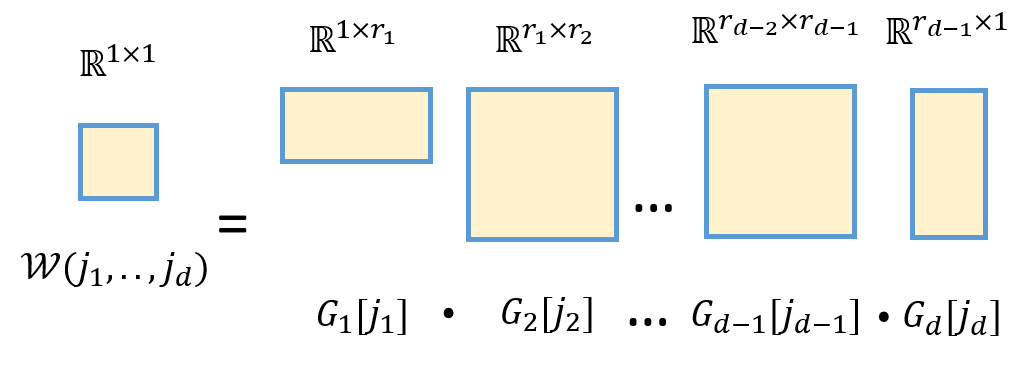}
    \caption{Illustration for Eq.\ref{eq:ttprod1}: Calculating an element $\mathcal{W}(j_1,..,j_k)$ using set of TT-cores $\{G_k[j_k]\}_{k=1}^{d}$}
    \label{fig:ttprodilus}
\end{figure}

By factoring the original tensor $\mathcal{W}$ into multiple TT-cores $\{\mathcal{G}_k\}_{k=1}^{d}$, we can compress the number of elements needed to represent the original tensor size from $\prod_{k=1}^d n_k$ to $\sum_{k=1}^{d} n_k r_{k-1}r_{k}$. 

\subsection{Representing Linear Transformation using TT-format}
Almost all of the parts of neural networks are composed of linear transformations: 
\begin{eqnarray} \label{eq:linearlayer}
y = W x + b,
\end{eqnarray} where $W \in \mathbb{R}^{M \times N}$ is the weight matrix and $b \in \mathbb{R}^{M}$ is the bias vector.
In most cases, matrix $W$ has many more parameters than bias $b$. Therefore, we can utilize the TT-format for optimizing our neural networks by replacing weight matrix $W$ with tensor $\mathcal{W}$ in TT-format \cite{novikov2015tensorizing}. 

We represent the TT-format for matrix $W \in \mathbb{R}^{M \times N}$ where $M=\prod_{k=1}^{d} m_k$ and $N=\prod_{k=1}^{d} n_k$ as tensor $\mathcal{W}$ by defining bijective functions $\mathbf{f}_i: \mathbb{Z}_{+} \rightarrow \mathbb{Z}_{+}^{d}$ and $\mathbf{f}_j : \mathbb{Z}_{+} \rightarrow \mathbb{Z}_{+}^{d}$. Function $\mathbf{f}_i$ maps each row $p \in \{1,..,M\}$ into $\mathbf{f}_i(p) = [i_1(p),..,i_d(p)]$ and $\mathbf{f}_j$ map each column $q \in \{1,..,N\}$ into $\mathbf{f}_j(q)=[j_1(q),..,j_d(q)]$. After defining such bijective functions, we can access the value from matrix $W(p, q)$ in tensor $\mathcal{W}$ with the index vectors generated by $\mathbf{f}_i(p)$ and $\mathbf{f}_j(q)$. We transform Eq.\ref{eq:ttprod1} to represent matrix $W$ in the TT-format:
\begin{eqnarray}
W(p, q) &=& \mathcal{W}(\mathbf{f}_i(p), \mathbf{f}_j(q)) \label{eq:ttmat1} \\
&=& \mathcal{W}\left(\left[i_1(p),..,i_d(p)\right], \left[j_1(q),..,j_d(q)\right]\right) \label{eq:ttmat2} \\
&=& G_1\left[i_1(p),j_1(q)\right] .. G_d\left[i_d(p),j_d(q)\right] \label{eq:ttmat3}
\end{eqnarray}
where for each $k \in \{1,..,d\}$:
\begin{eqnarray}
G_k[i_k(p), j_k(q)] &\in& \mathbb{R}^{r_{k-1} \times r_k} \nonumber \\
i_k(p) &\in& \{1,..,m_k\}\nonumber  \\
j_k(q) &\in& \{1,..,n_k\}\nonumber .
\end{eqnarray}
To represent the linear transformation in Eq.\ref{eq:linearlayer} with Eq.\ref{eq:ttmat1}-\ref{eq:ttmat3}, we need to reshape the vector input $x$ into tensor $\mathcal{X}$ and bias vector $b$ into tensor $\mathcal{B}$ with order $d$ to match our tensor $\mathcal{W}$. The following equation calculates a similar operation with $y(p) = W(p, :) x + b(p)$ where we map row index $p$ to vector $[i_{1}(p),..,i_{d}(p)]$ and enumerate all possible mappings for all columns in matrix $W$:
\begin{multline}
\mathcal{Y}\left(i_{1}(p),..,i_{d}(p)\right) =  \sum_{j_1,..,j_d} G_1[i_{1}(p),j_{1}] .. G_d[i_{d}(p),j_{d}] \cdot \\
\mathcal{X}\left(j_1,..,j_d\right) + \mathcal{B}\left(i_{1}(p),..,i_{d}(p)\right)
\end{multline}

\begin{table}[]
\centering
\caption{Fully Connected vs TT Layer Running Time and Memory}

\label{tbl:bigottlayer}
\begin{tabular}{|l|l|l|}

\hline
\multicolumn{1}{|c|}{\textbf{Operation}} & \multicolumn{1}{c|}{\textbf{Time}} & \multicolumn{1}{c|}{\textbf{Memory}} \\ \hline
FC forward & $O(MN)$ & $O(MN)$ \\ \hline
TT forward & $O(dr^2m \max(M,N))$ & $O(r \max(M,N)$)  \\ \hline
FC backward & $O(MN)$ & $O(MN)$ \\ \hline
TT backward & $O(d^2r^4m \max(M,N))$ & $O(r^3 \max(M,N))$                    \\ \hline
\end{tabular}
\end{table}

We can control the shape of TT-cores $ \{\mathcal{G}_k\}^{d}_{i=1} $ by choosing factor $M$ as $\{m_k\}_{k=1}^{d}$ and $N$ as $\{n_k\}_{k=1}^{d}$ as long as the number of factors is equal between $M$ and $N$. We can also define TT-rank $\{r_k\}_{k=0}^{d}$ and treat them as a hyper-parameter. In general, if we use a smaller TT-rank, we will get more efficient models but this action restricts our model to learn more complex representation. If we use a larger TT-rank, we get more flexibility to express our weight parameters but we sacrifice model efficiency. Table \ref{tbl:bigottlayer} compares the forward and backward propagation times and the memory complexity between the fully connected layer and the TT-layer in Big-O notation \cite{novikov2015tensorizing}. We compare the fully connected layer with matrix $W \in \mathbb{R}^{M \times N}$ versus the TT-layer with tensor $\mathcal{W}$ with TT-rank $\{r_k\}_{k=0}^{d}$. In the table, $m$ denotes $\max(\{m_k\}_{k=1}^{d})$, and $r$ denotes $\max(\{r_k\}_{k=0}^{d})$.

\subsection{Compressing Simple RNN with TT-format}
We represent a simple RNN in TT-format and call this model TT-SRNN for the rest of this paper. From Section \ref{sec:rnn}, we focus our attention on two dense weight matrices: ($W_{xh}, W_{hh}$). Previously, we defined $W_{xh} \in \mathbb{R}^{M \times N}$ as input-to-hidden parameters and $W_{hh} \in \mathbb{R}^{M \times M}$ as hidden-to-hidden parameters. 

First, we factorize matrix shape $M$ into $\prod_{k=1}^{d}m_k$ and $N$ into $\prod_{k=1}^{d}n_k$. Next, we determine TT-rank $\{r_k\}_{k=0}^{d}$ for our model and substitute $W_{xh}$ with tensor $\mathcal{W}_{xh}$ and $W_{hh}$ with tensor $\mathcal{W}_{hh}$. Tensor $\mathcal{W}_{xh}$ is represented by set of TT-cores $\{\mathcal{G}_k^{xh}\}_{k=1}^{d}$ where $\forall{k} \in \{1,..,d\}$, $\mathcal{G}_k^{xh} \in \mathbb{R}^{m_k \times n_k \times r_{k-1} \times r_k}$, and tensor $\mathcal{W}_{hh} $ is represented by set of TT-cores $\{\mathcal{G}_k^{hh}\}_{k=1}^{d}$ where $\forall{k} \in \{1,..,d\}$, $\mathcal{G}_k^{hh} \in \mathbb{R}^{m_k \times m_k \times r_{k-1} \times r_{k}}$. We define bijective functions $\mathbf{f}_i^{x}$ and $\mathbf{f}_i^{h}$ to access row $p$ from $W_{xh}$ and $W_{hh}$ in the set of TT-cores. We rewrite our simple RNN formulation to calculate $h_t$ in Eq.\ref{eq:simplernn}:
\begin{eqnarray}
a_{t}^{xh}(p) &=&  \sum_{j_1,..,j_d} \mathcal{W}_{xh}(\mathbf{f}_i^{x}(p), [j_1,..,j_d]) \cdot \mathcal{X}_t\left(j_1,..,j_d\right) \\
a_{t}^{hh}(p) &=&  \sum_{j_1,..,j_d} \mathcal{W}_{hh}(\mathbf{f}_i^{h}(p), [j_1,..,j_d]) \cdot \mathcal{H}_{t-1}\left(j_1,..,j_d\right) \\
a_{t}^{xh} &=& \left[a_t^{xh}(1),..,a_t^{xh}(M)\right] \\
a_{t}^{hh} &=& \left[a_t^{hh}(1),..,a_t^{hh}(M)\right] \\
h_t &=& f(a_t^{xh} + a_t^{hh} + b_h),
\end{eqnarray}
where $\mathcal{X}$ is the tensor representation of input $x_t$ and $\mathcal{H}_{t-1}$ is the tensor representation of previous hidden states $h_{t-1}$.

\subsection{Compressing GRU RNN with TT-format}
In this section, we apply TT-format to represent a gated RNN. Among several RNN architectures with gating mechanism, we choose GRU to be reformulated in TT-format because it has less complex formulation and similar performance as LSTM. We call this model TT-GRU for the rest of this paper. In Section \ref{sec:gru}, we focus on the following six dense weight matrices: ($W_{xr}$, $W_{hr}$, $W_{xz}$, $W_{hz}$, $W_{xh}$, and $W_{hh}$).  Weight matrices $W_{xr}$, $W_{xz}$, $W_{xh}$ $\in \mathbb{R}^{M \times N}$ are parameters for projecting the input layer to the reset gate, the update gate, the candidate hidden layer, and $W_{hr}$, $W_{hz}$, $W_{hh}$ $\in \mathbb{R}^{M \times M}$ are respectively parameters for projecting previous hidden layer into the reset gate, the update gate and candidate hidden layer.

We factorize $M=\prod_{k=1}^{d}m_k$, $N=\prod_{k=1}^{d}n_k$ and set TT-rank as $\{r_k\}_{k=0}^{d}$. All weight matrices ($W_{xr}$, $W_{hr}$, $W_{xz}$, $W_{hz}$, $W_{xh}$, $W_{hh}$) are substituted with tensors ($\mathcal{W}_{xr}$, $\mathcal{W}_{hr}$, $\mathcal{W}_{xz}$, $\mathcal{W}_{hz}$, $\mathcal{W}_{xh}$, $\mathcal{W}_{hh}$) in TT-format. Tensors $\mathcal{W}_{xr}$, $\mathcal{W}_{xz}$, $\mathcal{W}_{xh}$ are represented by a set of TT-cores ($\{\mathcal{G}_k^{xr}\}_{k=1}^d$, $\{\mathcal{G}_k^{xz}\}_{k=1}^d$, $\{\mathcal{G}_k^{xh}\}_{k=1}^d $) where $\forall{k} \in \{1,..,d\}, (\mathcal{G}_k^{xr}, \mathcal{G}_k^{xz}, \mathcal{G}_k^{xh} \in \mathbb{R}^{m_k \times n_k \times r_{k-1} \times r_k})$. Tensor $\mathcal{W}_{hr}$, $\mathcal{W}_{hz}$, $\mathcal{W}_{hh}$ are represented by a set of TT-cores ($\{\mathcal{G}_k^{hr}\}_{k=1}^d$, $\{\mathcal{G}_k^{hz}\}_{k=1}^d$, $\{\mathcal{G}_k^{hh}\}_{k=1}^d$) where $\forall{k} \in \{1,..,d\}$, $ (\mathcal{G}_k^{hr}, \mathcal{G}_k^{hz}, \mathcal{G}_k^{hh} \in \mathbb{R}^{m_k \times m_k \times r_{k-1} \times r_k})$. We define bijective function $\mathbf{f}_i^{x}$ to access row $p$ from $W_{xr}, W_{xz}, W_{xh}$ and function $\mathbf{f}_i^{h}$ to access row $p$ from $W_{hr}, W_{hz}, W_{hh}$ in the set of TT-cores. We rewrite the GRU formulation to calculate $r_t$ in Eq.\ref{eq:grureset}:
\begin{eqnarray}
a_{t}^{xr}(p) &=&  \sum_{j_1,..,j_d} \mathcal{W}_{xr}(\mathbf{f}_i^x(p), [j_1,..,j_d]) \cdot \mathcal{X}_t\left(j_1,..,j_d\right) \nonumber \\
a_{t}^{hr}(p) &=&  \sum_{j_1,..,j_d} \mathcal{W}_{hr}(\mathbf{f}_i^h(p), [j_1,..,j_d]) \cdot \mathcal{H}_{t-1}\left(j_1,..,j_d\right) \nonumber \\
a_{t}^{xr} &=& \left[a_t^{xr}(1),..,a_t^{xr}(M)\right] \nonumber \\
a_{t}^{hr} &=& \left[a_t^{hr}(1),..,a_t^{hr}(M)\right] \nonumber \\
r_t &=& \sigma(a_t^{xr} + a_t^{hr} + b_{r}).
\end{eqnarray}
Next, we rewrite the GRU formulation to calculate $z_t$ in Eq.\ref{eq:gruupdate}:
\begin{eqnarray}
a_{t}^{xz}(p) &=&  \sum_{j_1,..,j_d} \mathcal{W}_{xz}(\mathbf{f}_i^x(p), [j_1,..,j_d]) \cdot \mathcal{X}_t\left(j_1,..,j_d\right) \nonumber \\
a_{t}^{hz}(p) &=&  \sum_{j_1,..,j_d} \mathcal{W}_{hz}(\mathbf{f}_i^h(p), [j_1,..,j_d]) \cdot \mathcal{H}_{t-1}\left(j_1,..,j_d\right) \nonumber \\
a_{t}^{xz} &=& \left[a_t^{xz}(1),..,a_t^{xz}(M)\right] \nonumber \\
a_{t}^{hz} &=& \left[a_t^{hz}(1),..,a_t^{hz}(M)\right] \nonumber \\
z_t &=& \sigma(a_t^{xz} + a_t^{hz} + b_{z}).
\end{eqnarray}
Finally, we rewrite the GRU formulation to calculate $\tilde{h_t}$ in Eq.\ref{eq:grucandidate}: 
\begin{eqnarray}
a_{t}^{xh}(p) &=&  \sum_{j_1,..,j_d} \mathcal{W}_{xh}(\mathbf{f}_i^x(p), [j_1,..,j_d]) \cdot \mathcal{X}_t\left(j_1,..,j_d\right)\nonumber  \\
a_{t}^{hh}(p) &=&  \sum_{j_1,..,j_d} \mathcal{W}_{hh}(\mathbf{f}_i^h(p), [j_1,..,j_d]) \cdot \nonumber \\
& & \quad \left(\mathcal{R}_{t}\left(j_1,..,j_d\right) \cdot \mathcal{H}_{t-1}\left(j_1,..,j_d\right)\right) \nonumber \\
a_{t}^{xh} &=& \left[a_t^{xh}(1),..,a_t^{xh}(M)\right] \nonumber \\
a_{t}^{hh} &=& \left[a_t^{hh}(1),..,a_t^{hh}(M)\right] \nonumber \\
\tilde{h_t} &=& f(a_t^{xh} + a_t^{hh} + b_{h}).
\end{eqnarray}
After $r_t$, $z_t$ and $\tilde{h_t}$ are calculated, we calculate $h_t$ on Eq.\ref{eq:gruhidden} with standard operations like element-wise sum and multiplication. 

In practice, we could assign a different $d$ for each weight tensor as long as the input data dimension can also be factorized into the $d$ values. We could also put different TT-rank for each tensor and treat them as our model hyper-parameter. However, to simplify our implementation we use the same TT-rank for both the input and hidden projection weight tensors. We also use the same factorizations $M=\prod_{k=1}^{d} m_k$ and $N=\prod_{k=1}^{d} n_k$ for all weight tensors in TT-SRNN and TT-GRU.

We do not substitute bias vector $b$ into tensor $\mathcal{B}$ because the number of bias parameters is insignificant compared to the number of parameters in matrix $W$. In terms of performance, the element-wise sum operation for bias vector $b$ is also insignificant compared to the matrix multiplication between a weight matrix and the input layer or the previous hidden layer.

\subsection{Initialization for TT-cores Parameters} \label{sec:glorot}
Weight initialization is one critical detail for training deep neural networks. Especially for our RNN with TT-format that has many mini-tensors and several multiplications, the TT-RNN will have a longer matrix multiplication chain than a standard RNN, and the hidden layer value will quickly saturate \cite{glorot2010understanding}. Therefore, we need to carefully choose the initialization method to help our proposed model start in a stable condition. In our implementation, we follow Glorot initialization \cite{glorot2010understanding} to keep the same variance of weights gradients across layers and time-steps to avoid the vanishing gradient problem. We initialize all the TT-cores as follows:
\begin{eqnarray}
\forall k \in \{1,..,d\}, \quad \mathcal{G}_k &\sim& \mathcal{N}(0, \sigma_k),  \nonumber \\ 
\quad \text{where} \quad \sigma_k &=& \sqrt{\frac{2}{(n_k \cdot r_{k})+(m_k \cdot r_{k-1})}} \nonumber
\end{eqnarray}
By choosing a good initialization, our neural network will converge faster and obtain better local minima. Based on our preliminary experiments, we get better starting loss at the first several epochs compared to the randomly initialized model with the same $\sigma_k$ on Gaussian distribution for all TT-cores.

\section{Experiments} \label{sec:expr}
In this section, we evaluate our proposed RNN model with TT-formats (TT-SRNN and TT-GRU) and compare them to baseline RNNs (a simple RNN and GRU). We conducted the experiments on sequence classification tasks, where each input sequence was assigned a single class, and sequence prediction tasks, where we predicted the next time-step based on previous information \cite{graves2012supervised}. We used MNIST dataset for the sequence classification task and polyphonic music datasets for the sequence prediction task. For both tasks, we used local Glorot initialization trick from Section \ref{sec:glorot} for all the TT-cores weight parameters on the TT-SRNN and TT-GRU models. We used Adam algorithm \cite{kingma2014adam} to optimize our model parameters.

For reports on both tasks, we simplified the model description as follows: RNN-H$\bigstar $ where $\bigstar$ denotes the number of hidden units (e.g., RNN-H256 means RNN with 256 hidden units) and TT-SRNN-H$\bigstar$-R$\blacklozenge$ where $\blacklozenge$ denotes the TT-rank (e.g., TT-SRNN-H10x10-R3 means TT-SRNN with hidden units 10x10 in TT-format and TT-rank 3). We used a grid search to determine the best number of hidden layer units for both models and the shape of TT-format based on the validation set performance.

\subsection{Sequence Classification on Sequential MNIST}
We evaluated our proposed model TT-SRNN and TT-GRU for classification task using the MNIST dataset \cite{le2015simple}. The MNIST dataset consists of 28 x 28 grayscale images from ten classes (digits 0-9). The MNIST dataset has a training set with 50000 images, a development set with 10000 images, and a test set with 10000 images. We have three different ways to represent the MNIST dataset in our experiments.

For the first experiment, we fed each row starting at the top row and ending at the bottom row, which means we fed a vector with 28 values at each time-step and a total of 28 time-steps to represent an image. We used the latest hidden layer activation as our image representation and put a softmax layer to classify the digits. This task's difficulty is medium for a simple RNN and an easy task for gated RNN. Our baseline models consists of RNN and GRU with 256 hidden units. For our proposed model, we use TT-SRNN and TT-GRU with $10 \times 10$ shapes and ranks (3, 5). For all the models, we use a projection layer with 32 hidden units before we feed the input to our RNN. The projection layer is used to embed the pixel representation into richer feature representation. We show the result on Table \ref{tbl:mnistrow}. We repeated all of the experiments five times with different weight parameters initializations. Both the baseline and proposed models converged with good accuracy in several epochs and we achieved similar accuracy with a compression rate up to 80 times.

\begin{table}[]
\centering
\caption{Compression Rate and Accuracy for MNIST Row}
\label{tbl:mnistrow}
\begin{tabular}{|l|c|c|c|}
\hline
\multicolumn{1}{|c|}{\textbf{Model}} & \textbf{RNN Params} & \textbf{Compr.} & \textbf{Test Acc} \\ \hline
RNN-H256                             & 82176               & 1               & 96.5 $\pm$ 0.35             \\ \hline
TT-SRNN-H10x10-R3                      & 1030                & 79.78           & 96.9 $\pm$ 0.36     \\ \hline
TT-SRNN-H10x10-R5                      & 1700                & 48.34           & 97.1 $\pm$ 0.2             \\ \hhline{|=|=|=|=|}
GRU-H256                             & 221952              & 1               & 98.6 $\pm$ 0.12 \\ \hline
TT-GRU-H10x10-R3                      & 3180                & 69.8           & 98.3 $\pm$ 0.11 \\ \hline
TT-GRU-H10x10-R5                      & 5100                & 43.52           & 98.3 $\pm$ 0.12 \\ \hline
\end{tabular}
\end{table}

\begin{figure}[h]
    \centering
    \includegraphics[width=8.5cm]{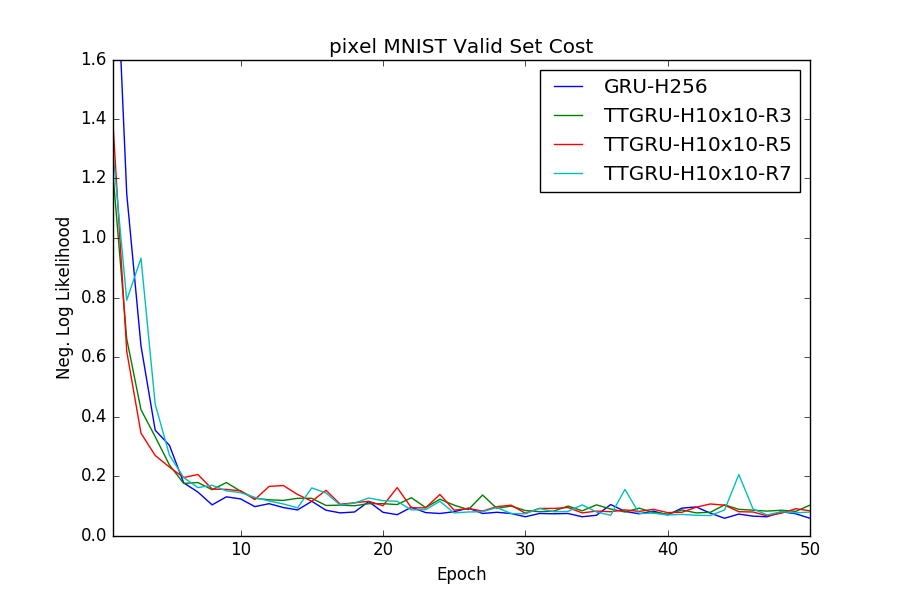}
    \caption{Comparison between baseline GRU with 256 hidden units, TT-GRU with $10 \times 10$ output shape and TT-rank ${3,5,7}$ on the pixel MNIST validation set.}
    \label{fig:mnistpixel}
\end{figure}

In our second experiment, we fed each pixel starting at the top left corner and ending at the bottom right corner, which means we fed a pixel at one time-step and in total we needed 784 time-steps to represent an image. This task is very challenging even for an RNN with gating mechanism because the RNN needs to model very long sequences \cite{le2015simple}. As in the first task, we fed the softmax layer using the latest hidden layer values. For this very long-dependency task, we only benchmarked the gated RNN variants (GRU and TT-GRU). For our proposed model, we used TT-GRU with output shapes (10, 10) and three different TT-ranks (3, 5, 7). For all the models, we use a projection layer with 32 hidden units before we feed the input to our RNN. Fig.~\ref{fig:mnistpixel} compares the validation set cost for each epoch. We can observe that the TT-GRU able to converge as fast as the baseline GRU model. In table \ref{tbl:mnistpixel}, our proposed model matched the baseline model with TT-rank 5 and reduced the parameters 43 times smaller compared to the baseline model.

\begin{table}[]
\centering
\caption{Compression Rate and Accuracy for pixel MNIST}
\label{tbl:mnistpixel}
\begin{tabular}{|l|c|c|c|}
\hline
\multicolumn{1}{|c|}{\textbf{Model}} & \textbf{RNN Params} & \textbf{Compr.} & \textbf{Test Acc} \\ \hline
GRU-H256                             & 221952              & 1               & 98.2              \\ \hline
TT-GRU-H10x10-R3                      & 3180                & 69.8            & 97.8              \\ \hline
TT-GRU-H10x10-R5                      & 5100                & 43.52           & 98.2              \\ \hline
TT-GRU-H10x10-R7                      & 7020                & 31.61           & 98.0              \\ \hline
\end{tabular}
\end{table}

\begin{figure}[h]
    \centering
    \includegraphics[width=8.5cm]{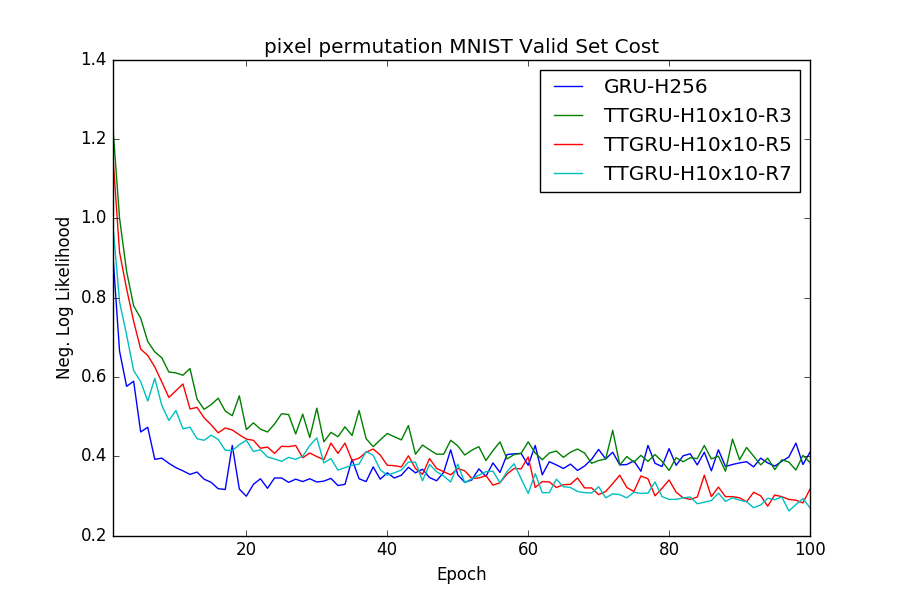}
    \caption{Comparison between baseline GRU with 256 hidden units, TT-GRU with $10 \times 10$ output shape and TT-rank ${3,5,7}$ on the p-MNIST validation set.}
    \label{fig:mnistpixelperm}
\end{figure}

In the last experiment, we used the most difficult task \cite{le2015simple} to push the limits of the gated RNN model. We shuffled the MNIST pixel-by-pixel and applied the same shuffled index to all the samples, and fed them one-by-one in a similar way as in the previous experiment. For the baseline and proposed model, we used the same configuration as in the previous experiment. Fig.~\ref{fig:mnistpixelperm} compares the validation set cost for each epoch. In Table \ref{tbl:mnistpixelperm}, we show that our proposed models was able to match the baseline models with TT-rank 5 and reduced the RNN parameters to 43 times smaller.

\begin{table}[]
\centering
\caption{Compression Rate and Accuracy for p-MNIST}
\label{tbl:mnistpixelperm}
\begin{tabular}{|l|c|c|c|}
\hline
\multicolumn{1}{|c|}{\textbf{Model}} & \textbf{RNN Params} & \textbf{Compr.} & \textbf{Test Acc} \\ \hline
GRU-H256                             & 221952              & 1               & 91.5              \\ \hline
TT-GRU-H10x10-R3                      & 3180                & 69.8            & 88.5              \\ \hline
TT-GRU-H10x10-R5                      & 5100                & 43.52           & 91.5              \\ \hline
TT-GRU-H10x10-R7                      & 7020                & 31.61           & 92.0              \\ \hline
\end{tabular}
\end{table}

\subsection{Sequence Prediction on Polyphonic Music}
\begin{table*}[t]
\centering
\caption{Compression Rate, Negative Log Likelihood and Accuracy for All Polyphonic Music Test Set}
\label{tbl:poly}
\begin{tabular}{|l|l|l|l|l|l|l|l|l|l|l|}
\hline
\multicolumn{1}{|c|}{\multirow{2}{*}{\textbf{Model}}} & \multicolumn{1}{c|}{\multirow{2}{*}{\textbf{Params}}} & \multicolumn{1}{c|}{\multirow{2}{*}{\textbf{Compr.}}} & \multicolumn{2}{c|}{\textbf{Nottingham}}                              & \multicolumn{2}{c|}{\textbf{PianoMidi}}                               & \multicolumn{2}{c|}{\textbf{MuseData}}                                & \multicolumn{2}{c|}{\textbf{JSB Chorales}}                            \\ \cline{4-11} 
\multicolumn{1}{|c|}{}                                & \multicolumn{1}{c|}{}                                     & \multicolumn{1}{c|}{}                                 & \multicolumn{1}{c|}{\textbf{NLL}} & \multicolumn{1}{c|}{\textbf{ACC}} & \multicolumn{1}{c|}{\textbf{NLL}} & \multicolumn{1}{c|}{\textbf{ACC}} & \multicolumn{1}{c|}{\textbf{NLL}} & \multicolumn{1}{c|}{\textbf{ACC}} & \multicolumn{1}{c|}{\textbf{NLL}} & \multicolumn{1}{c|}{\textbf{ACC}} \\ \hline
RNN-H512 & 393728 & 1                   & 3.45$\pm$0.04 & 70.4$\pm$0.4 & 7.66$\pm$0.02 & 26.9$\pm$0.1 & 7.31$\pm$0.03 & 35.6$\pm$0.2 & 8.41$\pm$0.04 & 29.3$\pm$0.2 \\ \hline
TT-SRNN-H8x4x8x4-R3 & 2560 & 153.80     & 3.59$\pm$0.03 & 69.5$\pm$0.3 & 7.72$\pm$0.04 & 27.8$\pm$0.4 & 7.69$\pm$0.02 & 32.9$\pm$0.4 & 8.56$\pm$0.05 & 28.8$\pm$0.3\\ \hline
TT-SRNN-H8x4x8x4-R5 & 4864 & 80.95      & 3.54$\pm$0.01 & 69.7$\pm$0.2 & 7.68$\pm$0.03 & 27.5$\pm$0.4 & 7.57$\pm$0.1 & 33.4$\pm$0.9 & 8.55$\pm$0.03 & 28.6$\pm$0.5 \\ \hhline{|=|=|=|=|=|=|=|=|=|=|=|}
GRU-H512 & 1181184 & 1                  & 3.35$\pm$0.02 & 71.4$\pm$0.1   & 7.59$\pm$0.01 & 26.7$\pm$0.4 & 7.12$\pm$0.02 & 36.4$\pm$0.7 & 8.32$\pm$0.01 & 30.6$\pm$0.3 \\ \hline
TT-GRU-H8x4x8x4-R3 & 7680 & 153.80      & 3.52$\pm$0.04 & 69.9$\pm$0.3 & 7.61$\pm$0.01 & 26.8$\pm$0.4 & 7.51$\pm$0.1 & 33.1$\pm$0.5 & 8.50$\pm$0.04 & 28.6$\pm$0.3 \\ \hline
TT-GRU-H8x4x8x4-R5 & 14592 & 80.95      & 3.48$\pm$0.04 & 70.4$\pm$0.3 & 7.59$\pm$0.01 & 27.5$\pm$0.2 & 7.44$\pm$0.15 & 35.0$\pm$1.0 & 8.48$\pm$0.02 & 28.5$\pm$0.3  \\ \hline
\end{tabular}
\end{table*}

For the sequential modeling tasks, we used four polyphonic music datasets \cite{boulanger2012}: Piano-midi.de, Nottingham, MuseData, and JSB Chorales. All of these datasets have 88 binary values per time-step, and each consists of at least seven hours of polyphonic music. Our baseline models are a simple RNN with 512 hidden units and a GRU RNN with 512 hidden units. Our proposed models are TT-SRNN and TT-GRU with $8\times4\times8\times4$ output shapes and TT-ranks (3, 5). Before we fed our input into the RNN, we projected them using hidden layer with 256 hidden units. In the polyphonic modeling task, we measured two different metrics: negative log-likelihood (NLL) and accuracy (ACC). To calculate the accuracy, we followed the evaluation metric proposed by \cite{bay2009evaluation} where $ACC = TP / (TP+FP+FN)$. We only used true positive (TP), false positive (FP), false negative (FN) and ignored the true negative (TN) because most of the notes were turned off in the dataset. Table \ref{tbl:poly} lists all of the results of our experiments on the baseline and proposed models. We repeat all experiments five times with different weight parameters initialization.

The table shows that all of these models have similar performances based on the negative log-likelihood and the accuracy in the test set. Our proposed models was able to reduce the number of parameter with significant compression ratio and preserved the performance at the same time.

\section{Related Work} \label{sec:relatedwork}
Compressing parameters on neural network architecture has become an interesting topic over the past several years due to the increased complexity of neural networks. The number of parameters and processing times has also grown tremendously along with their performance. A number of researchers comes up with many different ways to tackle this problem.

Ba et al. \cite{ba2014deep} and Hinton et al. \cite{hinton2015distilling} ``distilled" the knowledge from a deep neural network into a shallow neural network. First, they trained a state-of-the-art model with a deep and complex neural network using the original dataset and hard label as the target. After that, they reused the trained deep neural network by extracting output from the softmax layer and used them as the output target for a shallow neural network. By training the shallow network with a soft target, they achieved a better performance than the model trained using hard target labels. Recently, Tang et al. \cite{tang2016recurrent} utilized a similar approach for training RNN with a trained DNN. However, they had to train two different neural networks and built different structures to transfer the knowledge from bigger models.

From the probabilistic perspective, Graves et al. \cite{graves2011practical} proposed a variational inference method for learning the mean and variance of Gaussian distribution for each weight parameter. They reformulated the variational inference as the optimization of a Minimum Description Length \cite{hinton1993keeping}. By modeling each weight parameter, they learned the importance of each weight in regard to the model. After the training process was finished, they pruned the parameters by removing the weight that has a high probability to be zero. However, they still needed large matrix multiplication and represented their model in dense weight matrix, and thus the algorithmic and memory complexity remained the same as in the original model.

Another approach to tackle the compression problem by a technical perspective is to limit the precision for weight parameters. Gupta et al.\cite{gupta2015deep} and Courbariaux et al.\cite{courbariaux2014training} minimized the performance loss while using fewer bits (e.g., 16 bits) to represent floating points. Courbariaux et al.\cite{courbariaux2015binaryconnect} proposed BinaryConnect to constrain the weight possible values to -1 or +1. Most of these ideas can be easily applied with our proposed model since several deep-learning frameworks have built-in low-precision floating point options \cite{tensorflow2015-whitepaper, 2016arXiv160502688short}.

Model compression using low-rank matrix has also been reported \cite{denil2013predicting, sainath2013low}. Both of these works showed that many weight parameters are significantly redundant, and by representing them as low-rank matrices, they reduced the number of parameters with only a small drop in accuracy. Recently, Lu et al.\cite{lu2016learning} used low-rank matrix ideas to reduce the number of parameters in an RNN. Novikov et al.\cite{novikov2015tensorizing} utilized TT-format to represent weight matrices on feedforward neural networks. From their empirical evaluation on DNN-based architecture, the feedforward layer represented by the TT-format has a far better compression ratio and smaller accuracy loss compared to the low-rank matrix approach.

To the best of our knowledge, there are only a few research about compression on RNN models, and none of these works have utilized tensor-based format to represent the weight matrices for RNN models. In this work, we presented an RNN model by using TT-format weight to re-parameterize the weight matrices. We also compared the performance to standard uncompressed RNNs with a greater number of parameters. We expect our model could minimize the number of parameters and preserved the performance simultaneously.

\section{Conclusion} \label{sec:conclusion}
In this paper, we presented an efficient and compact RNN model using TT-format representation. Using TT-format, we represented dense weight matrices inside the RNN layer with multiple low-rank tensors. Our proposed TT-SRNN and TT-GRU significantly compressed the number of parameters while simultaneously retaining the model performance and accuracy. We evaluated our model with sequence classification and sequence prediction tasks. On sequence classification, with very long dependency tasks, our proposed RNNs model reduced the RNN parameters up to 40 times smaller compared to the original models without losing any accuracy. On the sequence prediction task, we evaluated our model with multiple music datasets, and our proposed RNNs reduced the RNN parameters up to 80 times smaller while preserving the performance.


\section*{Acknowledgment}
Part of this work was supported by Microsoft CORE 10 Project as well as JSPS KAKENHI Grant Numbers 24240032 and 26870371.



%
\bibliographystyle{IEEEtran}
\bibliography{refs}

\end{document}